\documentclass[conference]{IEEEtran}
\IEEEoverridecommandlockouts

\usepackage{cite}
\usepackage{amsmath,amssymb,amsfonts}
\usepackage{algorithmic}
\usepackage{graphicx}
\usepackage{textcomp}
\usepackage{xcolor}
\usepackage{multirow}
\usepackage{comment}
\usepackage{tabularx}
\usepackage{amsmath}
\usepackage{array}
\usepackage[colorlinks = true, citecolor = black, urlcolor = gray, linkcolor = black]{hyperref}

\begin{document}

\title{Analysis of Argument Structure Constructions in a Deep Recurrent Language Model}


\author{
Pegah Ramezani\textsuperscript{1,3}, 
Achim Schilling\textsuperscript{2,3}\textsuperscript{,}\textsuperscript{3}, 
Patrick Krauss\textsuperscript{3} \\
\textsuperscript{1}Department of English and American Studies, University Erlangen-Nuremberg, Germany\\
\textsuperscript{2}Neuroscience Lab, University Hospital Erlangen, Germany\\
\textsuperscript{3}Cognitive Computational Neuroscience Group, Pattern Recognition Lab, University Erlangen-Nuremberg, Germany\\
pegah.ramezani@fau.de, achim.schilling@fau.de, patrick.krauss@fau.de
}

\maketitle

\begin{abstract}
Understanding how language and linguistic constructions are processed in the brain is a fundamental question in cognitive computational neuroscience. In this study, we explore the representation and processing of Argument Structure Constructions (ASCs) in a recurrent neural language model. We trained a Long Short-Term Memory (LSTM) network on a custom-made dataset consisting of 2000 sentences, generated using GPT-4, representing four distinct ASCs: transitive, ditransitive, caused-motion, and resultative constructions.

We analyzed the internal activations of the LSTM model's hidden layers using Multidimensional Scaling (MDS) and t-Distributed Stochastic Neighbor Embedding (t-SNE) to visualize the sentence representations. The Generalized Discrimination Value (GDV) was calculated to quantify the degree of clustering within these representations. Our results show that sentence representations form distinct clusters corresponding to the four ASCs across all hidden layers, with the most pronounced clustering observed in the last hidden layer before the output layer. This indicates that even a relatively simple, brain-constrained recurrent neural network can effectively differentiate between various construction types.

These findings are consistent with previous studies demonstrating the emergence of word class and syntax rule representations in recurrent language models trained on next word prediction tasks. In future work, we aim to validate these results using larger language models and compare them with neuroimaging data obtained during continuous speech perception. This study highlights the potential of recurrent neural language models to mirror linguistic processing in the human brain, providing valuable insights into the computational and neural mechanisms underlying language understanding.
\end{abstract}



\section*{Introduction}

Understanding how language is processed and represented in the brain is a central challenge in cognitive neuroscience \cite{pulvermuller2002neuroscience}. In this paper, we adopt a usage-based constructionist approach to language, which views language as a system of form-meaning pairs (constructions) linking patterns to specific communicative functions \cite{goldberg2009nature, goldberg2003constructions}. In particular, argument Structure Constructions (ASCs) such as transitive, ditransitive, caused-motion, and resultative constructions play a crucial role in language comprehension and production \cite{goldberg1995constructions, goldberg2006constructions, goldberg2019explain}. These constructions are fundamental to syntactic theory and are integral to the way meaning is constructed in sentences. Investigating the neural and computational mechanisms underlying the processing of these constructions can provide significant insights into language and cognition \cite{pulvermuller2012meaning, pulvermuller2021biological, henningsen2022modelling, pulvermuller2023neurobiological}.

In recent years, advances in computational neuroscience have enabled the use of artificial neural networks to model various aspects of human cognition \cite{cohen2022recent}. Furthermore, the synergy between AI and cognitive neuroscience has led to a better understanding of the brain's unique complexities \cite{krauss2024artificial}. AI models, inspired by neural networks \cite{hassabis2017neuroscience}, have allowed neuroscientists to delve deeper into the brain's workings, offering insights that were previously unattainable \cite{krauss2023kunstliche}. These models have been particularly useful in studying how different parts of the brain interact and process information \cite{savage2019ai}. Among these neural network models, recurrent neural networks (RNNs) \cite{krauss2019weight, metzner2022dynamics, metzner2024quantifying}, and specifically Long Short-Term Memory (LSTM) networks \cite{hochreiter1997long}, have shown considerable promise in modeling sequential data, such as natural language \cite{wang2015learning}. Unlike transformer based large language models \cite{vaswani2017attention, radford2018improving}, which have gained popularity in natural language processing (NLP), LSTMs are designed to capture long-range dependencies in sequences without the need of a sliding window, making them more analogous to certain aspects of brain function related to temporal processing \cite{surendra2023word}.

This study employs a cognitive computational neuroscience approach \cite{kriegeskorte2018cognitive}. In particular, we explore how a deep recurrent language model, based on LSTM architecture, processes and represents different ASCs. We generated a custom dataset using GPT-4 \cite{radford2018improving, radford2019language;brown2020language}, comprising 2000 sentences evenly distributed across four ASC types. By training the LSTM model on this dataset for next word prediction, we aim to examine how well the model distinguishes between the different constructions at various levels of its internal representations.

To analyze the internal activations of the LSTM model, we utilized dimensionality reduction techniques such as Multidimensional Scaling (MDS) \cite{torgerson1952multidimensional} and t-Distributed Stochastic Neighbor Embedding (t-SNE) \cite{van2008visualizing} (cf. Methods). These techniques allow us to visualize high-dimensional data in a two-dimensional space, facilitating the identification of clusters corresponding to different ASCs. Additionally, we computed the Generalized Discrimination Value (GDV) \cite{schilling2021quantifying} to quantify the clustering quality, providing an objective measure of how well the model's internal representations align with the different construction types (cf. Methods).

Our findings indicate that the LSTM model successfully differentiates between the four ASC types, with the most distinct clustering observed in the final hidden layer before the output. This suggests that even a relatively simple, brain-constrained recurrent neural network can capture complex syntactic structures. These results are in line with previous research demonstrating the emergence of word class and syntactic rule representations in recurrent language models.

In future work, we plan to extend this research by validating our findings using large language models such as BERT \cite{devlin2018bert, krauss2024analyzing} and comparing the computational model's performance with neuroimaging data collected during continuous speech perception \cite{schilling2021analysis}. By bridging the gap between computational models and neural data, we aim to advance our understanding of the neural mechanisms underlying language processing \cite{kriegeskorte2018cognitive}.

This study highlights the potential of recurrent neural language models to mirror linguistic processing in the human brain, offering valuable insights into the computational and neural mechanisms that underpin language understanding.


\section*{Methods}

\subsection*{Dataset creation using GPT4}

To investigate the processing and representation of different Argument Structure Constructions (ASCs) in a recurrent neural language model, we created a custom dataset using GPT-4. This dataset was designed to include sentences that exemplify four distinct ASCs: transitive, ditransitive, caused-motion, and resultative constructions (cf. Table \ref{tab:agree_that_construction}). Each ASC category consisted of 500 sentences, resulting in a total of 2000 sentences.

\subsubsection*{Selection of Argument Structure Constructions}

The four ASCs selected for this study are foundational to syntactic theory and represent different types of sentence structures: \\
Transitive Constructions: Sentences where a subject performs an action on a direct object (e.g., "The cat chased the mouse"). \\
Ditransitive Constructions: Sentences where a subject performs an action involving a direct object and an indirect object (e.g., "She gave him a book"). \\
Caused-motion Constructions: Sentences where a subject causes an object to move in a particular manner (e.g., "He pushed the cart into the garage"). \\
Resultative Constructions: Sentences where an action results in a change of state of the object (e.g., "She painted the wall red").

\begin{table}[h]
\centering
\begin{tabular}{|p{0.20\linewidth}|p{0.28\linewidth}|p{0.28\linewidth}|}
\hline
\textbf{Constructions} & \textbf{Structure} & \textbf{Example} \\
\hline
Transitive & Subject + Verb + Object & The baker baked a cake. \\
\hline
Ditransitive & Subject + Verb + Object1 + Object2 & The teacher gave students homework. \\
\hline
Caused-Motion & Subject + Verb + Object + Path & The cat chased the mouse into the garden. \\
\hline
Resultative & Subject + Verb + Object + State & The chef cut the cake into slices. \\
\hline
\end{tabular}
\caption{Name, structure, and example of each construction}
\label{tab:agree_that_construction}
\end{table}

\subsubsection*{Generation of Sentences}

To ensure the diversity and quality of the sentences in our dataset, we utilized GPT-4, a state-of-the-art language model developed by OpenAI \cite{}. The generation process involved the following steps:
Prompt Design: We created specific prompts for GPT-4 to generate sentences for each ASC category. These prompts included example sentences and detailed descriptions of the desired sentence structures to guide the model in generating appropriate constructions.
Sentence Generation: Using the designed prompts, we generated 500 sentences for each ASC category. The generation process was carefully monitored to ensure that the sentences adhered to the syntactic patterns of their respective constructions.
Manual Review and Filtering: After the initial generation, we manually reviewed the sentences to ensure their grammatical correctness and adherence to the intended ASC types. Sentences that did not meet these criteria were discarded and replaced with newly generated ones.
Balancing the Dataset: To prevent any bias in the model training, we ensured that the dataset was balanced, with an equal number of sentences (500) for each of the four ASC categories.

\subsubsection*{Handling Varying Sentence Lengths}

Sentences in natural language vary in length, which poses a challenge for processing within neural networks. To address this, we used padding to standardize sentence lengths. Specifically, each sentence was padded to match the length of the longest sentence in the dataset. This padding ensures that all input sequences are of equal length, facilitating efficient batch processing during model training.

\subsubsection*{Text Tokenization}

To convert the textual data into a numerical format suitable for input into the neural network, we used a tokenizer. The tokenization process involved the following steps:
Vocabulary Creation: Each unique word in the dataset was identified and assigned a specific ID number. This process resulted in a vocabulary list where each word corresponded to a unique integer identifier.
Sentence Transformation: Each sentence was transformed into a sequence of these integer IDs, representing the words in the order they appeared. For instance, a sentence like "The cat chased the mouse" would be converted into a sequence of integers based on the IDs assigned to each word.
By padding sentences to a uniform length and converting them into numerical sequences, we ensured that the dataset was ready for training the LSTM-based recurrent neural language model. These preprocessing steps are crucial for enabling the model to effectively learn and differentiate between the various ASCs.

Using word IDs instead of word embeddings in this study offers several advantages. Firstly, word IDs provide a simpler and more interpretable representation of the dataset, which aligns well with the study's focus on analyzing internal model activations and clustering of sentence representations based on Argument Structure Constructions (ASCs). This simplicity aids in isolating the effects of syntactic structures without the added complexity of pre-trained embeddings that carry semantic information from external contexts. Secondly, using word IDs ensures that the analysis remains focused on the syntactic and structural aspects of sentence processing, allowing for a clearer examination of how the LSTM model differentiates between different ASCs. This approach facilitates a more straightforward interpretation of the model's ability to capture syntactic patterns, which is the primary interest of this research.

The resulting dataset, comprising 2000 sentences represented as padded numerical sequences, serves as a robust foundation for training and analyzing the LSTM model. This carefully curated and preprocessed dataset enables us to investigate how different ASCs are processed and represented within the model, providing insights into the underlying computational mechanisms.

\subsection*{LSTM architecture}

The LSTM model in this study is designed for next-word prediction without prior information about the type of sentence constructions. The initial goal is to evaluate the model's ability to predict the next word, while the main objective is to assess how well it understands and differentiates between the different constructions. The model architecture consists of four layers:
Embedding Layer: This layer converts each sentence into a sequence of integer numbers, transforming the input words into dense vector representations. This step facilitates efficient processing by the LSTM layers.
First LSTM Layer: This layer learns complex patterns and dependencies within the sequence of words, capturing the contextual information necessary for accurate next-word prediction.
Second LSTM Layer: Building upon the first LSTM layer, this layer further refines the learned patterns and dependencies, enhancing the model's understanding of the sequence's structure.
Dense Layer with Softmax Activation: The final layer outputs a probability distribution over all possible next words. The softmax activation function ensures that the output is a valid probability distribution, suitable for predicting the next word.
The model ultimately outputs a one-hot vector, where the length corresponds to the number of possible next words, indicating the predicted probabilities for each word. This architecture enables the model to learn and represent the intricate patterns of different Argument Structure Constructions (ASCs), providing insights into how such constructions are processed and differentiated by a recurrent neural language model.

\subsection*{Analysis of Hidden Layer Activations}

After training the model, we assessed its ability to differentiate between the various constructions by analyzing the activations of its hidden layers. Given the high dimensionality of these activations, direct visual inspection is not feasible. To address this, we employed dimensionality reduction techniques to project the high-dimensional activations into a two-dimensional space.
By combining different visualization and quantitative techniques, we were able to assess the model's internal representations and its ability to differentiate between the various linguistic constructions.

\subsubsection*{Multidimensional Scaling (MDS)}

This technique was used to reduce the dimensionality of the hidden layer activations, preserving the pairwise distances between points as much as possible in the lower-dimensional space. In particular, MDS is an efficient embedding technique to visualize high-dimensional point clouds by projecting them onto a 2-dimensional plane. Furthermore, MDS has the decisive advantage that it is parameter-free and all mutual distances of the points are preserved, thereby conserving both the global and local structure of the underlying data \cite{torgerson1952multidimensional, kruskal1964nonmetric,kruskal1978multidimensional,cox2008multidimensional, metzner2021sleep, metzner2023extracting, metzner2022classification}. 

When interpreting patterns as points in high-dimensional space and dissimilarities between patterns as distances between corresponding points, MDS is an elegant method to visualize high-dimensional data. By color-coding each projected data point of a data set according to its label, the representation of the data can be visualized as a set of point clusters. For instance, MDS has already been applied to visualize for instance word class distributions of different linguistic corpora \cite{schilling2021analysis}, hidden layer representations (embeddings) of artificial neural networks \cite{schilling2021quantifying,krauss2021analysis}, structure and dynamics of highly recurrent neural networks \cite{krauss2019analysis, krauss2019recurrence, krauss2019weight, metzner2023quantifying}, or brain activity patterns assessed during e.g. pure tone or speech perception \cite{krauss2018statistical,schilling2021analysis}, or even during sleep \cite{krauss2018analysis,traxdorf2019microstructure,metzner2022classification,metzner2023extracting}. 
In all these cases the apparent compactness and mutual overlap of the point clusters permits a qualitative assessment of how well the different classes separate.

\subsubsection*{t-Distributed Stochastic Neighbor Embedding (t-SNE)}

This method further helped in visualizing the complex structures within the activations by emphasizing local similarities, allowing us to see the formation of clusters corresponding to different Argument Structure Constructions (ASCs). t-SNE is a frequently used method to generate low-dimensional embeddings of high-dimensional data  \cite{van2008-visualizing}. However, in t-SNE the resulting low-dimensional projections can be highly dependent on the detailed parameter settings \cite{wattenberg2016use}, sensitive to noise, and may not preserve, but rather often scramble the global structure in data \cite{vallejos2019exploring, moon2019visualizing}. Here, we set the perplexity (number of next neighbours taken into account) to 100.

\subsection*{Generalized Discrimination Value (GDV)}

To quantify the degree of clustering, we used the GDV as published and explained in detail in \cite{schilling2021quantifying}. This GDV provides an objective measure of how well the hidden layer activations cluster according to the ASC types, offering insights into the model's internal representations. Briefly, we consider $N$ points $\mathbf{x_{n=1..N}}=(x_{n,1},\cdots,x_{n,D})$, distributed within $D$-dimensional space. A label $l_n$ assigns each point to one of $L$ distinct classes $C_{l=1..L}$. In order to become invariant against scaling and translation, each dimension is separately z-scored and, for later convenience, multiplied with $\frac{1}{2}$:
\begin{align}
s_{n,d}=\frac{1}{2}\cdot\frac{x_{n,d}-\mu_d}{\sigma_d}.
\end{align}
Here, $\mu_d=\frac{1}{N}\sum_{n=1}^{N}x_{n,d}\;$ denotes the mean,\\ \\
and $\sigma_d=\sqrt{\frac{1}{N}\sum_{n=1}^{N}(x_{n,d}-\mu_d)^2}$ the standard deviation of dimension $d$. \\ \\
Based on the re-scaled data points $\mathbf{s_n}=(s_{n,1},\cdots,s_{n,D})$, we calculate the {\em mean intra-class distances} for each class $C_l$ 
\begin{align}
\bar{d}(C_l)=\frac{2}{N_l (N_l\!-\!1)}\sum_{i=1}^{N_l-1}\sum_{j=i+1}^{N_l}{d(\textbf{s}_{i}^{(l)},\textbf{s}_{j}^{(l)})},
\end{align}
and the {\em mean inter-class distances} for each pair of classes $C_l$ and $C_m$
\begin{align}
\bar{d}(C_l,C_m)=\frac{1}{N_l  N_m}\sum_{i=1}^{N_l}\sum_{j=1}^{N_m}{d(\textbf{s}_{i}^{(l)},\textbf{s}_{j}^{(m)})}.
\end{align}
Here, $N_k$ is the number of points in class $k$, and $\textbf{s}_{i}^{(k)}$ is the $i^{th}$ point of class $k$.
The quantity $d(\textbf{a},\textbf{b})$ is the euclidean distance between $\textbf{a}$ and $\textbf{b}$. Finally, the Generalized Discrimination Value (GDV) is calculated from the mean intra-class and inter-class distances  as follows:
\begin{align}
\mbox{GDV}=\frac{1}{\sqrt{D}}\left[\frac{1}{L}\sum_{l=1}^L{\bar{d}(C_l)}\;-\;\frac{2}{L(L\!-\!1)}\sum_{l=1}^{L-1}\sum_{m=l+1}^{L}\bar{d}(C_l,C_m)\right]
 \label{GDVEq}
\end{align}

\noindent whereas the factor $\frac{1}{\sqrt{D}}$ is introduced for dimensionality invariance of the GDV with $D$ as the number of dimensions.

\vspace{0.2cm}\noindent Note that the GDV is invariant with respect to a global scaling or shifting of the data (due to the z-scoring), and also invariant with respect to a permutation of the components in the $N$-dimensional data vectors (because the euclidean distance measure has this symmetry). The GDV is zero for completely overlapping, non-separated clusters, and it becomes more negative as the separation increases. A GDV of -1 signifies already a very strong separation.

\subsection*{Code implementation, Computational resources, and programming libraries}

All simulations were run on a standard personal computer. The evaluation software was based on Python 3.9.13 \cite{oliphant2007python}. For matrix operations the numpy-library \cite{van2011numpy} was used and data visualization was done using matplotlib \cite{hunter2007matplotlib} and the seaborn library \cite{waskom2021seaborn}. The dimensionality reduction through MDS and t-SNE was done using the sci-kit learn library.

The models were coded in Python. The neural networks were designed using the Keras \cite{keras} and Keras-RL \cite{kerasrl} libraries. Mathematical operations were performed with numpy \cite{numpy} and scikit-learn \cite{scikit-learn} libraries.
Visualizations were realized with matplotlib \cite{matplot} and networkX \cite{networkX}. For natural language processing we used SpaCy \cite{explosion2017spacy}.


\section*{Results}

To understand how the LSTM model differentiates between various Argument Structure Constructions (ASCs), we visualized the activations of its hidden layers using Multidimensional Scaling (MDS) and t-Distributed Stochastic Neighbor Embedding (t-SNE). Additionally, we quantified the degree of clustering using the Generalized Discrimination Value (GDV).

\begin{figure}[t]
  \centering
  \includegraphics[width=\columnwidth]{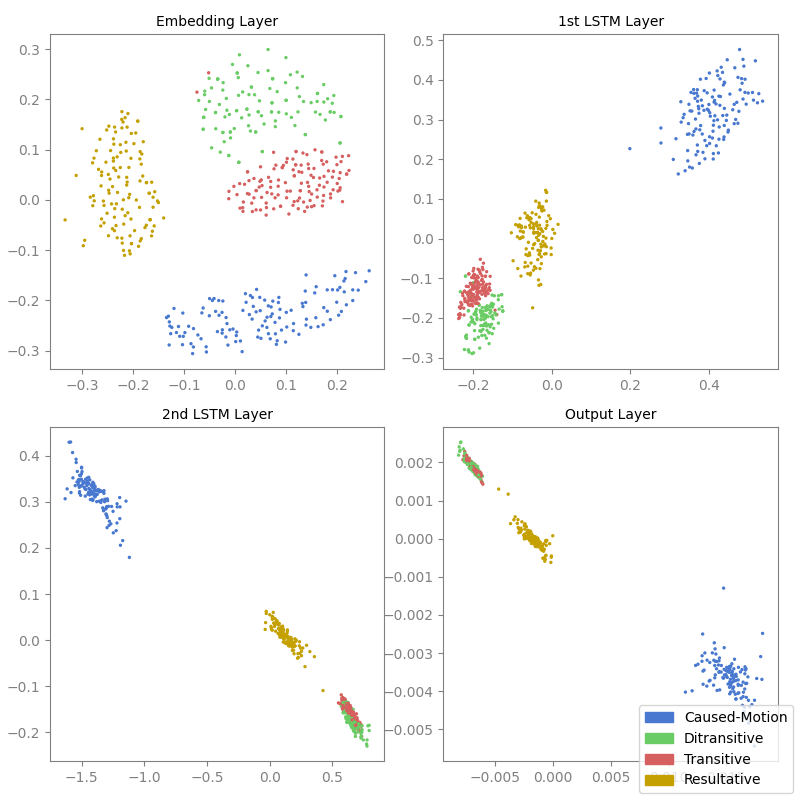}
  \caption{MDS projections of the activations from all four layers of the LSTM model. Each point represents the activation of a sentence, color-coded according to its ASC type: caused-motion (blue), ditransitive (green), transitive (red), and resultative (orange).}
  \label{fig:MDS}
\end{figure}

Figure \ref{fig:MDS} shows the MDS projections of the activations from all four layers of the LSTM model. Each point represents the activation of a sentence. The initial hidden layer already shows some separation between the different ASC types. As we move to the second LSTM layer, the separation between ASC types becomes more apparent, particularly with respect to the inter-cluster distances. However, the clusters for transitive and ditransitive sentences are closer to each other. In the third layer, the inter-cluster distances further increase, while the clusters for transitive and ditransitive sentences remain close to each other, indicating that the model is learning to differentiate between the ASCs more effectively and recognizes the similarity between transitive and ditransitive sentences. In the final output layer, the degree of clustering decreases slightly.

\begin{figure}[t]
  \centering
  \includegraphics[width=\columnwidth]{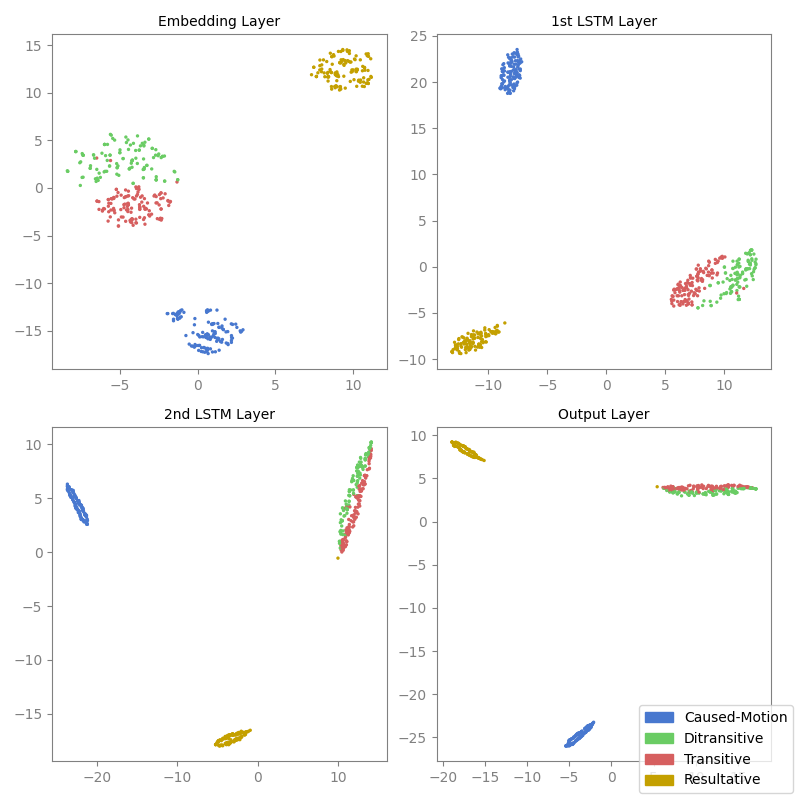}
  \caption{t-SNE projections of the activations from all four layers of the LSTM model. Each point represents the activation of a sentence, color-coded according to its ASC type: caused-motion (blue), ditransitive (green), transitive (red), and resultative (orange).}
  \label{fig:tSNE}
\end{figure}

The corresponding t-SNE projections shown in Figure \ref{fig:tSNE} yield qualitatively very similar results. The initial hidden layer shows some separation between ASC types, with increased and more apparent separation in the second layer, particularly in inter-cluster distances; this separation continues to improve in the third layer, while transitive and ditransitive sentences remain similar. The final layer shows a slight decrease in clustering degree.

\begin{figure}[t]
  \centering
  \includegraphics[width=\columnwidth]{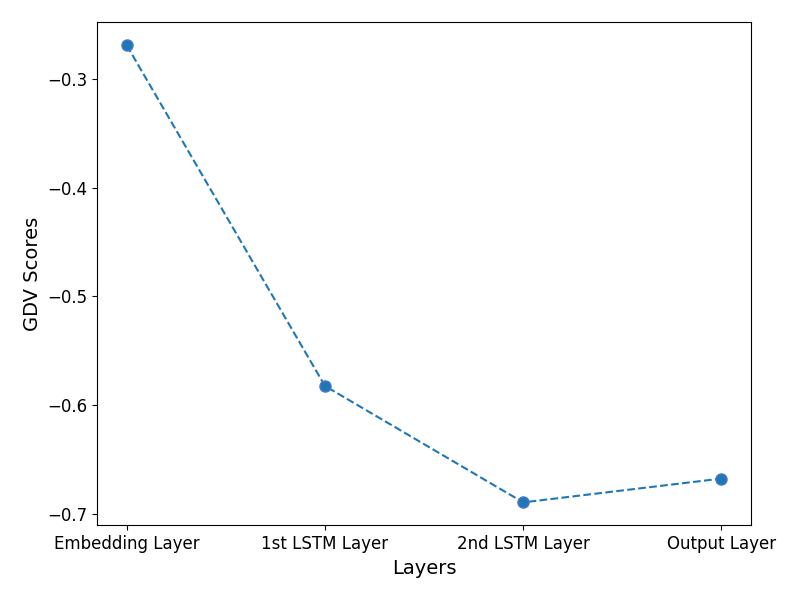}
  \caption{GDV score of hidden layer activations. Note that, lower GDV values indicate better-defined clusters. The qualitative results from the MDS and t-SNE projections are underpinned by the GDV with best clustering occurring in layer 3.}
  \label{fig:GDV}
\end{figure}

To quantitatively assess the clustering quality, we calculated the GDV for the activations of each hidden layer (cf. Figure \ref{fig:GDV}). Lower GDV values indicate better defined clusters. The qualitative results of the MDS and t-SNE projections are supported by the GDV. 

\section*{Discussion}

Our study aimed to understand how a recurrent neural language model (RNN) processes and represents different Argument Structure Constructions (ASCs) through the lens of cognitive computational neuroscience. Using a custom-generated dataset of sentences exemplifying four ASC types—transitive, ditransitive, caused-motion, and resultative—we trained an LSTM-based model for next-word prediction. The internal activations of the model’s hidden layers were analyzed and visualized using Multidimensional Scaling (MDS) and t-Distributed Stochastic Neighbor Embedding (t-SNE), with clustering quality quantified by the Generalized Discrimination Value (GDV).

Our analysis revealed that the model’s sentence representations formed distinct clusters corresponding to the four ASCs in all hidden layers. This indicates the model’s ability to differentiate between various syntactic structures. The clustering was most pronounced in the final hidden layer, just before the output layer. This suggests that as the information progresses through the layers, the model refines its understanding and separation of different ASCs.

The emergence of distinct ASC representations in our LSTM model aligns with previous studies that observed the formation of word class and syntax rule representations in recurrent language models trained on next-word prediction tasks \cite{surendra2023word}. This consistency across studies reinforces the idea that even relatively simple, brain-constrained neural network architectures \cite{pulvermuller2023neurobiological} like LSTMs can capture complex syntactic structures inherent in natural language.

Our findings suggest that recurrent neural networks can serve as effective computational analogs for studying linguistic processing in the human brain. The ability of the LSTM model to differentiate between ASCs supports the notion that similar neural mechanisms might be at play in human language comprehension.

The pronounced clustering in the final hidden layer hints at a hierarchical processing structure, where initial layers capture basic features, and subsequent layers integrate and refine these features into more complex representations. This parallels theories of hierarchical processing in the human brain \cite{golestani2014brain, badcock2019hierarchically, raut2020hierarchical}.

\subsection*{Limitations and Future work}

Our custom dataset, while carefully generated and balanced, is limited to 2000 sentences and four specific ASCs. Future studies could expand the dataset to include a wider variety of constructions and larger sentence pools to ensure generalizability.

Furthermore, our model used word IDs instead of embeddings, focusing on syntactic structures without semantic information. Incorporating pre-trained word embeddings \cite{almeida2019word} in future studies could provide a more holistic view of how semantic and syntactic information interact in neural representations.

While our computational findings are promising, they need to be validated against empirical neuroimaging data. Comparing the LSTM’s internal representations with brain activation patterns during continuous speech perception \cite{schilling2021analysis, schuller2023attentional, garibyan2022neural} could provide deeper insights into the neural correlates of ASC processing. Techniques like EEG and MEG could be used to collect neural data during language tasks \cite{schuller2024early}, enabling a direct comparison with the model’s activations using techniques such as representational similarity analysis \cite{kriegeskorte2008representational}. This would help bridge the gap between computational models and real-world brain function \cite{meeter2007neural, kriegeskorte2018cognitive}.

Although LSTMs are effective, they represent an earlier generation of neural network architectures \cite{hochreiter1997long}. Exploring more advanced models, such as transformers \cite{vaswani2017attention}, could provide additional insights into the processing and representation of ASCs \cite{goldberg1995constructions, goldberg2006constructions}. Transformers, with their attention mechanisms, might offer a more nuanced understanding of how different constructions are represented and processed, potentially revealing more about the interaction between different levels of linguistic information.

\section*{Conclusion}

Our study demonstrates that even a relatively simple LSTM-based recurrent neural network can effectively differentiate between various Argument Structure Constructions, mirroring some aspects of human linguistic processing. The distinct clustering of sentence representations suggests that the model captures essential syntactic structures, supporting its use as a computational tool in cognitive neuroscience. Future work should aim to validate these findings with larger datasets and neuroimaging data, and explore the capabilities of more advanced neural network architectures. By doing so, we can further our understanding of the computational and neural mechanisms underlying cognition and language processing in brains, minds and machines \cite{tuckute2024language, schilling2023auditory}.


\section*{Author contributions}
All authors discussed the results and approved the final version of the manuscript.

\section*{Acknowledgements}
This work was funded by the Deutsche Forschungsgemeinschaft (DFG, German Research Foundation): KR\,5148/3-1 (project number 510395418), KR\,5148/5-1 (project number 542747151), and GRK\,2839 (project number 468527017) to PK, and grant SCHI\,1482/3-1 (project number 451810794) to AS. 


\end{document}